\documentclass[runningheads]{llncs}
\usepackage{graphicx}
\usepackage{hyperref}
\usepackage{xcolor}

\begin{document}
\title{EPANer Team Description Paper\\ for World Robot Challenge 2020\thanks{Team introduction video:~\url{https://youtu.be/W38BwhI34d4}}}
%
%
\author{Zhi Yan\inst{1}
\and
Nathan Crombez\inst{1}
\and
Li Sun\inst{2}
}
\authorrunning{Z. Yan et al.}
%
\institute{
    University of Technology of Belfort-Montbéliard (UTBM), France
    \email{firstname.lastname@utbm.fr}\\
    \and
    University of Sheffield, UK\\
    \email{li.sun@sheffield.ac.uk}
}
\maketitle              
\begin{abstract}
This paper presents the research focus and ideas incorporated in the EPANer robotics team, entering the World Robot Challenge 2020 - Partner Robot Challenge (Real Space).

\keywords{World Robot Summit 2020 \and Toyota HSR robot \and Active perception \and Mobile grasping \and ROS.}
\end{abstract}
%
%
%
%
\section{Introduction}
\label{sec:Introduction}

EPANer\footnote{\url{https://epan-utbm.github.io/epaner/}} is a team that represents the EPAN Research Group at the University of Technology of Belfort-Montb\'eliard (UTBM) in France, founded in January 2018, aiming to participate to robotics competitions. Current members include:
\begin{itemize}
\item Dr. Zhi Yan\footnote{\url{https://yzrobot.github.io/}}, Assistant Professor, founder and team leader, mainly responsible for robot navigation, human tracking, and system integration.
\item Dr. Li Sun\footnote{\url{https://sites.google.com/site/lisunspersonalsite/}}, Assistant Professor, co-founder, mainly responsible for semantic mapping and mobile grasping.
\item Fahad Lateef, Ph.D. student, joined in January 2018, mainly responsible for camera-based scene understanding.
\item Yassine Ruichek, Professor, joined in January 2018 as team advisor.
\item Toyota HSR robot No. 88, joined in August 2018.
\item Dr. Nathan Crombez\footnote{\url{http://nathancrombez.free.fr/}}, Assistant Professor, joined in September 2018, mainly responsible for object detection and mobile grasping.
\item Thomas Duvinage, undergraduate student, joined in May 2019, works with Dr. Yan in robot navigation.
\item Pierig Servouze, undergraduate student, joined in May 2019, works with Dr. Crombez in mobile grasping.
\end{itemize}

EPANer participated in the World Robot Challenge 2018 - Partner Robot Challenge (Real Space) in Tokyo in October 2018, and eventually took the fifth place among 14 teams from various countries.
Some mature methods and conventional ideas used in robotics were quickly integrated into the competition on-site, such as finite-state machine for task flow control, waypoint-based topological navigation for robot navigation, 3D point cloud processing for object pose estimation, and YOLO algorithm for object recognition.
On the other hand, the team also learned some lessons and gained some research inspiration from the competition.

After returning to France, EPANer started to focus on the research aspects of active perception and mobile grasping with the HSR robot, and in the meantime, constantly improved the integrated system applicable to the competition.
In the following, we will introduce some of our ideas and technical details.
%
%
%
\section{Technical Challenge}
\label{sec:Technical Challenge}

The Service Category Technical Committee (hereinafter referred to as ``the Committee'') of the World Robot Challenge 2020 proposed the 4S philosophy for service robot performance benchmarking, including speed, smooth/smart, stable, and safe~\cite{DetailedRules}.
The following is our understanding in these regards.

\subsection{Speed}
Given a task that a robot needs to accomplish, the \emph{speed} metric means that the robot should complete the task in the shortest possible time~\cite{yz15iros,yz12phd}.
To do so, on the one hand, the robot essentially needs to reasonably decompose the task and properly plan the sub-tasks after decomposition~\cite{yz13ijars}.
On the other hand, the robot should move along the shortest planned path and avoid dynamic objects during the move.
For the latter, human-aware~\cite{kruse13survey} and socially compliant~\cite{kretzschmar16ijrr} mobile robot navigation are hopeful, instead of simply ``stopping-and-going''.
Furthermore, if the robot can quickly and accurately learn and reason about the environment around it, properly model and predict uncertainty, and even build a spatio-temporal model in a long-term work~\cite{vintr19icra,fremen}, then there will be a significant room for improvement in task completion speed.
For the spatio-temporal modeling, a typical example is that the robot is expected to clean the room at noon, because people are usually in the dining room for lunch at that time and thus allowing the robot to work quietly.

\subsection{Smooth/Smart}

\emph{Smooth} or \emph{smart} can be examined in several ways.
First, same as the \emph{speed} metric, efficient robot task and motion planning would contribute to this point, while the smoothness of motion can also benefit from the hardware design such as the omnidirectional mobile base of the HSR robot.
Bormann \emph{et al.}~\cite{bormann15icra} presented a prototype for office cleaning robot, in which the SMACH\footnote{\url{http://wiki.ros.org/executive_smach}} hierarchical state machine is used for smooth cleaning task control.
Palmieri and Arras~\cite{palmieri14iros} introduced a RRT (Rapidly-exploring Random Tree) extend function for efficient and smooth mobile robot motion planning.
Second, environment reasoning and learning is not only suitable for \emph{speed}, but also for \emph{smooth} and \emph{smart} purposes.
More personalized prediction like user preferences~\cite{abdo15icra} is more in line with the needs of the latter.
Third, high-level map representation including semantic~\cite{ls18ral,zhao17icar} and time~\cite{vintr19icra,fremen} information is shown to be more advanced and promising.

\subsection{Stable}

We prefer to interpret the \emph{stable} metric from the perspective of algorithm robustness and software engineering, respectively.
For the former, we are for example particularly attentive to develop a stable grasping process.
Visual servoing seems to be the appropriate approach to produce a \emph{smooth} and also \emph{stable} objects grasping and doors opening.
Using Photometric Gaussian Mixtures~\cite{crombez19tro} as visual features brings many advantages.
First, it does not require any feature detection, matching, or tracking process.
The redundancy of visual information provides a highly accurate and large convergence to drive the robot hand to the desired grasping pose.
This dense visual feature has shown to be particularly robust to image noise, occlusions and environment lighting changes.
Moreover, we are aiming to mount a Intel Realsense RGB-D camera on the HSR hand to include depth information in our control loop in order to increase our grasping success ratio.

On the other hand, software engineering such as unit testing, version control, and continuous integration is also an important method for stabilizing robot software systems.
Our previous work focused on building a ROS-based open source tested for multiple mobile robot systems\footnote{\url{https://github.com/yzrobot/mrs_testbed}}, while defined a benchmarking process based on experimental designs~\cite{yz17robotics}, which aimed at improving the reproducibility of experiments by making explicit all elements of a benchmark such as parameters, measurements and metrics.
This effort is also towards the direction of system stability.

\subsection{Safe}

Robot safety is an eternal topic that runs through the development of robotics and is not even limited to the discussion of technology itself.
The Three Laws of Robotics explain that \emph{safe} is not only for humans, but also for the robot itself.
For the former, human detection and tracking~\cite{nb18eor} is the key.
Shackleford \emph{et al.}~\cite{shackleford16jint} provided a method and a set of performance metrics that allow users to decide whether they can safely implement their application using human detection sensors.
The ongoing EU-funded ILIAD project\footnote{\url{https://iliad-project.eu/}} is a pioneer research project on intra-logistics with AGVs (Automated Guided Vehicle), which takes human safe as the core idea.
Under this framework, Mansfeld \emph{et al.}~\cite{mansfeld18ral} proposed the \emph{safety map} concept, that serves as a common unified representation for injury biomechanics data and robot collision behavior.
As for the safety of the robot itself, it can also be interpreted from two perspectives of algorithm and software engineering.
For the former, the classic obstacle avoidance algorithm implicitly considers the protection of the robot itself, while more advanced artificial intelligence like by modelling the epistemic uncertainty of deep policy network~\cite{sun2020icra} allows us to look forward to the robot's further response to ``dangerous'' such as collisions and ill-posed configurations.
From the software engineering perspective, one of the core evolutions of ROS2 over ROS is its security.
ROS2 will be built upon the DDS (Data Distribution System) specified by OMG (Object Management Group), which actually protects the robot from external program attacks while it is running.

\subsection{Discussion}

It actually can be seen that for service robots, in order to meet the 4S standard, in addition to addressing the related problems mentioned above, there is a very critical point that needs to be done, that is human detection, tracking, and motion trajectory prediction~\cite{rudenko19survey}.
EPANer has done a lot of work in these regards.
Specifically, we proposed an online learning framework for human detection and tracking experimented with different sensors~\cite{majer19ecmr,yz19auro,yz18iros,yz17iros}.
We also developed a 3DOF pedestrian trajectory prediction approach learned from long-term autonomous mobile robot deployment data with deep neural networks~\cite{ls18icra}.


Time- and site- specific activities have been less investigated in the competition. With consideration of the periodical change of the human activities \cite{hypertime,vintr19icra,fremen}, the conventional mapping can be extended to a fourth dimension, i.e. \emph{time}, thereby facilitating the task and motion planning and decision making for the socially-compliment navigation.
For example, the operator is allowed to indicate the time when entering an instruction to the robot, such as ``please bring this beer to Tom, it is 8pm now." and the robot should first go to the dining room to find Tom instead of going to another place.
%
%
%
\section{System Design}
\label{sec:System Design}

The Committee advocates the integration of ``keep moving'', ``move carefully'' and ``be clever'' concepts into robot system design for future challenges.
Here are some of our efforts and ideas that fit these three proposals.

\subsection{Keep moving}
We mainly address this point from three aspects, namely reasonable task decomposition and planning, reliable object and human detection, tracking, and trajectory prediction, and rapid online recovery from failures with less human intervention.
For the first aspect, we use the popular ROS SMACH package.
For the second aspect, we currently use 3D point cloud processing and rely on YOLO for object classification.
For human detection and tracking, we use a combination of RGB-D camera based upper body detector~\cite{jafari14icra} and 2D lidar based leg detector~\cite{arras07icra} for human detection, with a Bayesian filtering approach for human tracking~\cite{BayesianTracking}.
Our previous research~\cite{ls18icra} on human trajectory prediction cannot be integrated into the current system without modification, as the model is trained from long-term observation data based in a specific site and is unlikely to be generalized to other sites.
The same difficulty appeared in the adaption of the deep learned semantic understanding and mapping methods~\cite{sun2019weakly,zhao17icar} - the model needs to be retained with the new environment.
For the last aspect, we intend to adopt an approach similar to the behavior-performance map~\cite{cully15nature}.
When the system is restarted, the robot first returns to the last normal working state, and then takes an alternative from a predefined solution pool to continue the task.

\subsection{Move carefully}
In this regard, we believe that it is common to the \emph{keep moving} concept in terms of object and human detection, tracking, and trajectory prediction.
In addition, because we emphasize \emph{carefully} in this concept, some safety-efficiency trade-off  must be made.
For example, ``stop-and-go'' would be more secure and socially compliant in some cases, rather than trying to meet humans, especially in the case of robot holding objects.
This example also introduces another reference factor for \emph{carefully}, i.e. whether the robot holds an object and what object it holds. Obviously, holding a cup of coffee should be more careful than usual.
  
\subsection{Be clever}
One of the main directions of our research is the online learning method~\cite{majer19ecmr,yz19auro,yz18iros,yz17iros}, which fits well with the idea of making robots \emph{clever}.
This method is not only suitable for fast self-learning and environmental adaptation, but also for online recovery via short-term learning.
For the latter, errors, noises and outliers are able to be detected online, and then to be avoided in the next learning iterations.
Moreover, the error model can even be learned online for future error detection. 
Other than that, as we mentioned before, \emph{time} is also an important factor in making robots smart.
Learning the characteristics and laws of environmental changes from the long-term deployment will greatly improve the performance of service robots.

\section{Software Development}
\label{sec:Software Development}

Our framework has been fully implemented into the Robot Operating System (ROS) with high modularity.
C++ and Python programming languages are both used.
All members' development records are committed to a dedicated private repository on GitHub for collaborative development and built by Travis CI for continuous integration\footnote{According to the NDA agreement with Toyota, we do not compile the source code online with the TMC library.}.
The Docker container is not yet integrated.
However, building a continuous integration platform based on ``GitHub + Docker + Travis CI'' is our ultimate goal.

\begin{figure}
   \includegraphics[width=\textwidth]{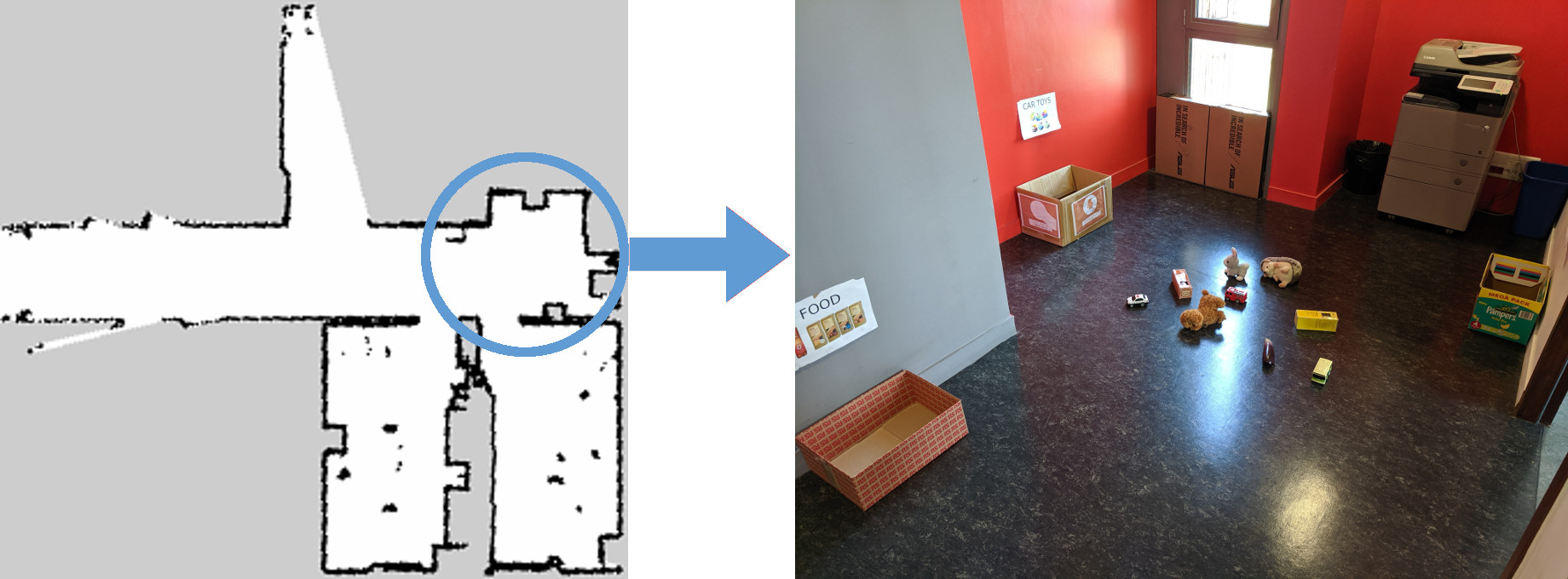}
   \caption{Our current arena for experimenting tidy-up tasks.}
   \label{fig:lab-map}
\end{figure}

The open source software we have used so far is as follows: joystick\_remapper\footnote{\url{https://github.com/epan-utbm/joystick_remapper}}, waypoint\_generator\footnote{\url{https://github.com/epan-utbm/waypoint_generator}}, executive\_smach\footnote{\url{https://github.com/ros/executive_smach}},
ViSP\footnote{\url{https://github.com/lagadic/vision_visp}},
find-object\footnote{\url{https://github.com/introlab/find-object}},
FLOBOT\footnote{\url{https://github.com/LCAS/FLOBOT}},
and YOLO\footnote{\url{https://pjreddie.com/darknet/yolo/}}.
The role of each member in software development is mentioned in the Sec.~\ref{sec:Introduction}.
Fig.~\ref{fig:lab-map} shows our experimental scenario.

%
%
%
%
\section{Relevant publications}

Our relevant publications include~\cite{nb18eor,crombez19tro,sun2020icra,majer19ecmr,ls18icra,ls18ral,sun2019weakly,vintr19icra,yz12phd,yz19auro,yz17iros,yz15iros,yz17robotics,yz13ijars,yz18iros,zhao17icar}.
%
%
%
%
%
%
%
\section{Conclusions}

In this paper, we presented the research focus and ideas incorporated in the EPANer robotics team, entering the World Robot Challenge 2020 - Partner Robot Challenge (Real Space).
We believe that the challenge is a very good way to benchmark our research results, as well as a best practice for our students and also a festival where peers can learn and communicate with each other.
Based on our results of World Robot Challenge 2018, we are more ambitious this time and look forward to continuing to participate in 2020.
\section*{Acknowledgments}
We thank Prof. Tom{\'{a}}s Krajn{\'{\i}}k\footnote{\url{http://labe.felk.cvut.cz/~tkrajnik/}} for helpful suggestions regarding ``time in robotics'' and his team for sharing ideas and giving comments.
We also thank NVIDIA Corporation for donating high-power GPUs on which part of our work was performed.
%
%
%
\bibliographystyle{splncs04}
\bibliography{mybibliography}
\label{references}

\end{document}